\documentclass{ETHpaper}

\usepackage{latexsym}

\usepackage[T1]{fontenc}
\usepackage[utf8]{inputenc}

\usepackage{microtype}
\usepackage{amsmath}
\usepackage{multirow}
\usepackage{booktabs}

\usepackage{mathtools}
\usepackage{tabularx}

\usepackage{caption}
\usepackage{enumitem}
\usepackage{fontawesome}

\usepackage{tabularx,booktabs,dcolumn, longtable, tabu}
\newcolumntype{Y}{>{\centering\arraybackslash}X}
\usepackage{stfloats}
\usepackage{makecell}

\definecolor{colorSG}{RGB}{168,50,45}
\definecolor{customY}{HTML}{FBB13C}
\definecolor{customG}{HTML}{218380}
\definecolor{customT}{HTML}{73D2DE}
\definecolor{customW}{HTML}{FCFCFF}
\definecolor{customB}{HTML}{2E5EAA}
\definecolor{customP}{HTML}{5B4E77}
\definecolor{gitRedFaded}{HTML}{ffeef0}
\definecolor{gitGreenFaded}{HTML}{e6ffed}
\definecolor{gitRed}{HTML}{ffdce0}
\definecolor{gitGreen}{HTML}{cdffd8}
\definecolor{gitRedFull}{HTML}{cb2431}
\definecolor{gitGreenFull}{HTML}{2cbe4e}
\definecolor{densityP}{HTML}{6600CC}
\definecolor{densityO}{HTML}{FF9933}
\definecolor{partyB}{HTML}{0000FF}
\definecolor{partyR}{HTML}{F73131}

\usepackage{cleveref}
\usepackage{graphicx}

\begin{document}

\title{Disentangling Active and Passive Cosponsorship in the U.S. Congress}

\author{
\begin{tabularx}{.7\textwidth}{CC}
Giuseppe Russo$^{*}$ & Christoph Gote$^{*}$ \\
\normalfont\itshape russog@ethz.ch & \normalfont\itshape cgote@ethz.ch\\[3mm]
Laurence Brandenberger$^{*}$ & Sophia Schlosser$^{*}$\\
\normalfont\itshape lbrandenberger@ethz.ch & \normalfont\itshape sschlosser@ethz.ch
\end{tabularx}\\[3mm]
\begin{tabularx}{.7\textwidth}{C}
Frank Schweitzer$^{*}$\\
\normalfont\itshape fschweitzer@ethz.ch
\end{tabularx}\\[3mm]
\normalfont\footnotesize\itshape
$^{*}$Chair of Systems Design, ETH Zurich, Weinbergstrasse 56/58, 8092 Zurich, Switzerland
}

\authoralternative{Russo, et al.}
\www{\url{http://www.sg.ethz.ch}}
\reference{Version of: \today}

\date{May 2021}

\maketitle

\begin{abstract}
    In the U.S. Congress, legislators can use active and passive cosponsorship to support bills.
    We show that these two types of cosponsorship are driven by two different motivations: the backing of political colleagues and the backing of the bill's content.
    To this end, we develop an Encoder+RGCN based model that learns legislator representations from bill texts and speech transcripts. 
    These representations predict active and passive cosponsorship with an F1-score of 0.88.
    Applying our representations to predict voting decisions, we show that they are interpretable and generalize to unseen tasks.
\end{abstract}

\section{Introduction}\label{sec:Introduction}

Expressing political support through the cosponsorship of bills is essential for the proper execution of congressional activities.
In the US Congress, legislators can draft bills and introduce them to the congress floor, after which they are referred to a committee for assessment. 
Once a legislative draft passes the committee, it is discussed in the plenary.
Here, legislators defend their stance and debate the bill's merits.
Finally, a bill is voted on.
Throughout the entire process---from a bills' conception until the final vote---legislators have the possibility to cosponsor the bill. 

The act of cosponsoring a bill is a strong signal of support of a legislator towards that bill's stance on a political topic.  
Legislators use these signals to understand which political topics are relevant right now and how their colleagues position themselves on the topic.
Thus, cosponsorship has been extensively studied.
Some studies examine the signaling power of cosponsorship signatures  \cite{kessler1996dynamics, wilson1997cosponsorship} and their effect on the passing of a bill at the final vote \cite{browne1985multiple, woon2008bill, sciarini2021influence, Dockendorff2021cosponsorship}. 
Other studies focus on understanding alliance dynamics between legislators \cite{fowler2006connecting, kirkland2011relational, kirkland2014measurement, lee2017time}.

\begin{figure}
    \centering
    \includegraphics[width=.5\linewidth]{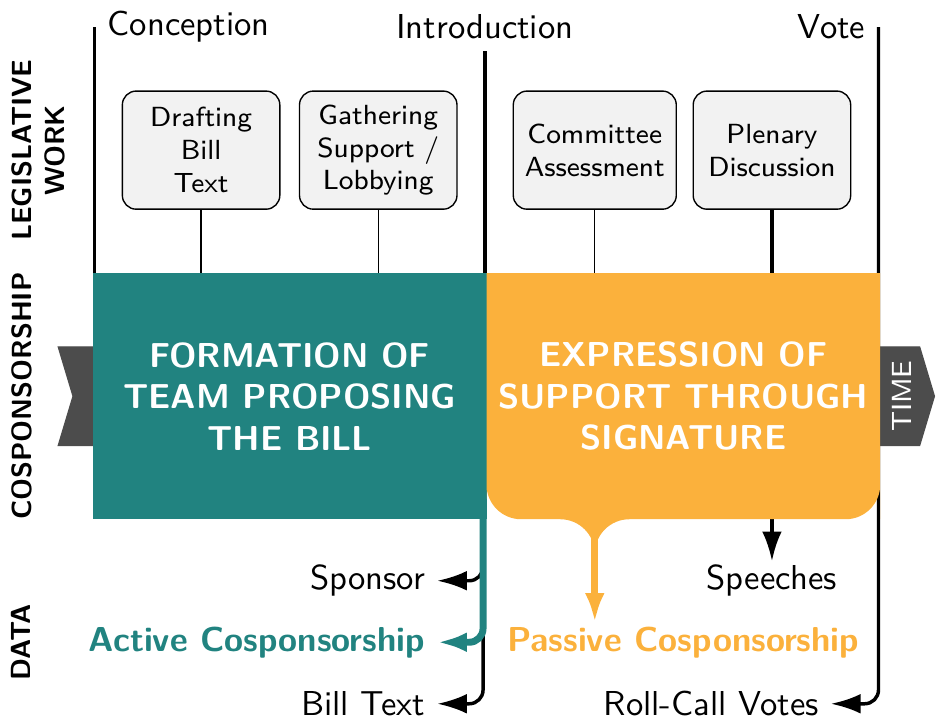}
    \caption{Distinction between active and passive cosponsorship in time and their relation to the legislative work.  
    Active cosponsorship {\textcolor{customG}{\footnotesize\faSquare}} occurs at the initial phase, before the bill is introduced.
    Passive cosponsorship {\textcolor{customY}{\footnotesize\faSquare}} occurs during the deliberation phase of the bill.
    }
    \label{fig:billprocess}
\end{figure}

In the US Congress, cosponsorship is differentiated between \textit{active} and \textit{passive}. 
As illustrated in \Cref{fig:billprocess}, the timing of cosponsorship determines this differentiation.
Active cosponsorship entails involvement---together with the legislator introducing the bill (\emph{sponsor})---in the bill’s creation in its initial stages.
In contrast, passive cosponsorship is be issued after the introduction of a bill to the Congress floor.

So far, most studies analyzing cosponsorship have not differentiated between active and passive cosponsorship.
These two actions have been distinguished with respect to their effort required.
Active cosponsorship can be considered as a more resource-intense form of support, given that legislators can be involved in the drafting process of a bill and help gather support.
In turn, passive cosponsorship is viewed as less resource-intense with a minimal effort to sign the bill \cite{fowler2006connecting}.
However, no studies so far have examined the underlying motivations that drive a legislator to actively or passively cosponsor a bill.
Given the importance of cosponsorship as a signal of support for a bill during a legislative process, we believe that it is crucial to understand not only why a legislator cosponsors a bill, but to understand why a legislator opts for an active or a passive cosponsorship.

\begin{figure*}
    \centering
    \includegraphics[width=\textwidth]{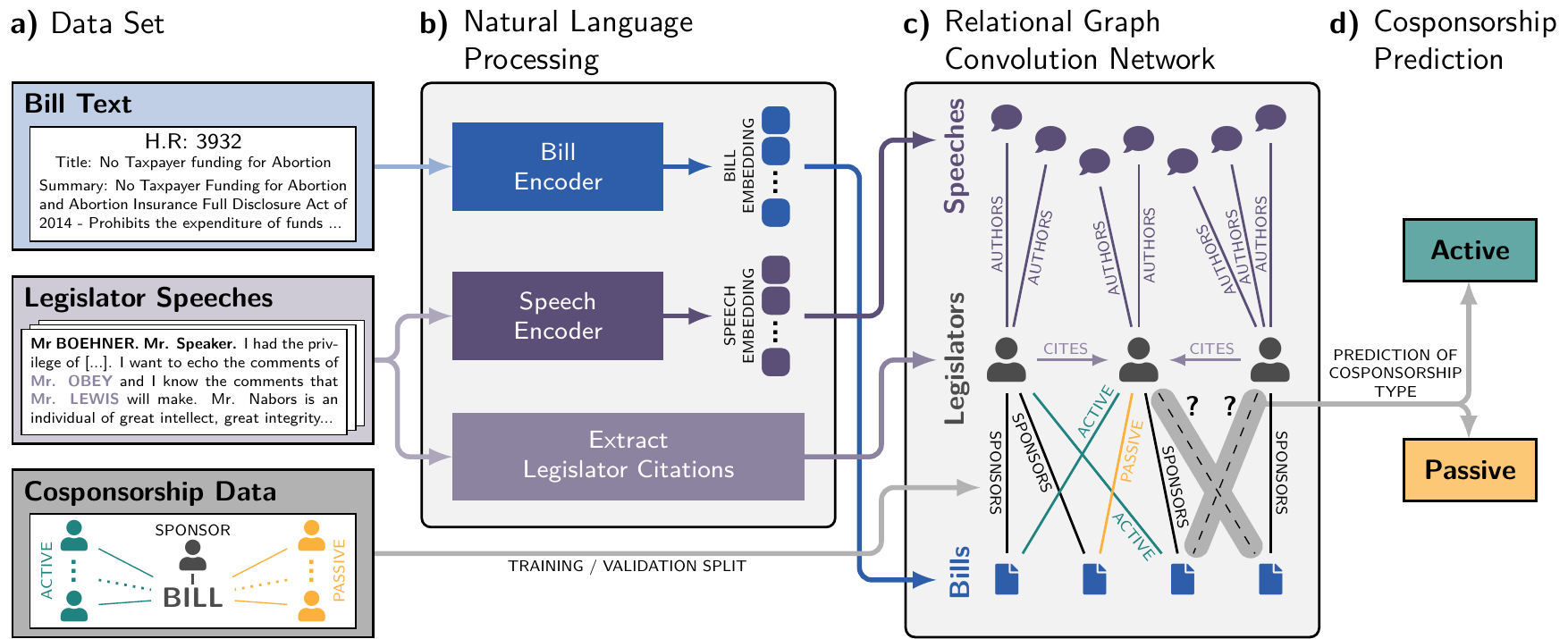}
    \caption{Overview of our Model.
      \textbf{a)} Our data contains bill texts, legislator speeches, and cosponsorship data for all bills from the 112th to 115th U.S. Congress.
      \textbf{b)} We use Natural language processing to obtain contextual embeddings of bills and speeches and to extract a citation network between legislators.
      \textbf{c)} We develop a Relational Graph Convolutional Network (RGCN) is trained based on a subset of the cosponsorship relations.
      \textbf{d)} The trained RGCN predicts active and passive cosponsorship relations in the validation and test data.} %
    \label{fig:infographic}
\end{figure*}

This work shows that active and passive cosponsorship is driven by two different motivations.
Active cosponsorship is people-centric and primarily signals the backing of the \emph{sponsor} of the bill.
In contrast, passive cosponsorship is driven by backing a bill's \emph{content}.
We summarize our approach to obtain these insights in \Cref{fig:infographic}.
Our contributions are as follows:
\begin{itemize}[label=\raisebox{.05em}{\tiny\faPlay}]
    \item We create a curated data set containing information on all bills and speeches from the 112th to 115th U.S. Congress.
    \item We develop a BERT+LSTM based encoder \citep{hochreiter1997long}. 
    Acknowledging that written and spoken language have different characteristics, we separately process bill texts and transcripts of legislator speeches to obtain embeddings capturing their content.
    \item We propose a Relational Graph Convolutional Network (RGCN) that learns legislator representations accounting for (i) the speeches they give, (ii) the bills they sponsor and cosponsor, and (iii) the other legislators they cite in their speeches.
    \item In a binary classification task, we use these representations to predict active/passive cosponsorship with an F1-score of 88\%. 
    Our representations show that active cosponsorship should be interpreted as a backing of the sponsor of a bill.
    Passive cosponsorship should be interpreted as a backing of the content of a bill.
    \item Our representations can be used as proxies for legislators' ideology. 
    Specifically, we show that they separate legislators matching both party and caucus memberships.
    In addition, they match task-specific state-of-the-art models for voting prediction without requiring additional training. 
    Hence, our legislator representations are interpretable and generalize to unseen tasks.
\end{itemize}

\section{Data}\label{sec:Data}

For our study, we collect fine-grained data on all bills and legislators from the 112th to 115th U.S. Congress.
Our data set contains (i) metadata for all legislators, (ii) bill texts, (iii) transcripts of all speeches mapped to the corresponding legislator, (iv) disambiguated data capturing which legislators sponsored and actively or passively cosponsored each bill, and (v) the resulting roll-call votes for all bills.
We provide detailed summary statistics for our data set \Cref{sec:AppendixData}.

\paragraph{Legislator Metadata}
We obtain the BioGuide ID, first name, last name, gender, age, party affiliation, state, and district of all legislators from \url{voteview.com}, a curated database containing basic data related to the U.S. Congress.

\paragraph{Bill Text}
As mentioned above, legislators introduce bills to propose laws or amend existing ones in order to further their agenda.
We acquire IDs, titles, and introduction dates of bills using the API of \url{propublica.org}, a non-profit organization that collects and provides access to congressional documents.
We further collect summaries of the bill's content, which the API provides for around 95\% of all cases.
For bills where no summary is available, we use the full-body texts instead.
As we create our data set to study active and passive cosponsorship, we discard all bills for which no cosponsorship links were recorded.
Overall, our data set contains information on over $50,000$ bills.

\paragraph{Legislator Speeches}
Legislators take the floor to advocate or oppose bills.
In these speeches, they communicate their agenda to their colleagues in order to persuade them to vote for (or against) a bill.
We obtain transcripts of congressional speeches by scraping \url{congress.gov}, the official website of the U.S. Congress.
The transcripts are archived in so-called \emph{daily editions}, which are effectively concatenations of all speeches from a day written verbatim.
All congressional speeches start with a formal introduction of the legislator giving the speech and the session's chairperson, e.g., ``Mr. POE of Texas. Mrs. President.'' or ``Mr. BOEHNER. Mr. Speaker'' (cf. \Cref{fig:infographic}a).
Using this pattern, we can split the daily editions and recover the individual speeches and speakers as follows:
First, we tag names and geopolitical entities (e.g., ``of Texas'') using the Named Entity Recognition model from SpaCy\footnote{\url{spacy.io/api/entityrecognizer}} with \texttt{[PERSON]} and \texttt{[GPE]} tags, respectively.
Second, we tag all salutations (e.g., Mrs/Mr) and institutional roles (e.g., Speaker, President) with \texttt{[SAL]} and \texttt{[ROLE]}. 
In doing so, the start of speeches is tagged either as \texttt{[SAL]+[PERSON]+[SAL]+[ROLE]} or \texttt{[SAL]+[PERSON]+[GPE]+[SAL]+[ROLE]}.
The \texttt{[PERSON]} tag further identifies the legislator giving the speech.\\
With this simple procedure, we map roughly 93\% of the speeches to the correct legislator.
We perform manual data cleaning on the speeches excluding subsets for three reasons:
(i) Speeches for which we cannot determine an author are predominantly given by a legislator representing a committee or an office. 
When legislators speak on behalf of an office or committee, the opinion expressed in the speech not necessarily corresponds to their personal opinion.
(ii) We found many speeches with less than 10 sentences that only contain procedural information.
(iii) Similarly, very long speeches with more than 500 sentences are usually of a commemorative nature, paying tribute to or praising a person, an institution, or an event. 
Both (ii) and (iii) convey no information on the legislators' stances.
Excluding these speeches from our data set, we obtain a total of over $120,000$ speech transcripts.\\
Finally, as shown in \Cref{fig:infographic}a, legislators frequently cite each other in speeches.
To detect citations in a speech, we first collect all entities that SpaCy tags as \texttt{[PERSON]}.
To distinguish instances in which speeches cite other legislators compared to third parties, we utilize the fact that in daily editions, the names of legislators are always written in upper case.
We match the names of legislators to their BioGuide IDs, resulting in a citation network.

\paragraph{Cosponsorship Data}
We identify the sponsor of all bills using the API of \url{propublica.org}.
In addition, the API provides the names of the legislators who cosponsored a bill and when this cosponsorship occurred.
We automatically match the cosponsors' names to their BioGuide ID.
In cases where automated matching was not possible---e.g., because legislators signed with their nicknames---we resorted to manual matching.
As discussed in \Cref{sec:Introduction}, we assign cosponsorship their official label. 
Cospsonsorships recorded at the bill's introduction are \emph{active}, and those recorded after its introduction are \emph{passive}.

\paragraph{Roll-Call Votes}
Roll-call votes are records of how legislators voted on bills.
We scrape these data using the Python package of \citet{pujari-goldwasser-2021-understanding}, yielding over $1.5$ million votes, which we match to the corresponding legislator and bill IDs.

\section{Methodology}\label{sec:Methods}
Our model to classify cosponsorship decisions based on the legislator and bill data described in the previous section consists of two main elements---an Encoder and a Relational Graph Convolutional Network (RGCN).
The Encoder computes high dimensional representations of legislators' bills and speeches based on their texts and transcripts, respectively. 
These representations are used by an RGCN and a downstream Feed-Forward Neural Network (FFNN) allowing us to predict how (i.e., active or passive) a cosponsor supports a bill.

\subsection{Encoder}\label{sec:Encoder}

\begin{figure}
    \centering
    \includegraphics[width=.5\linewidth]{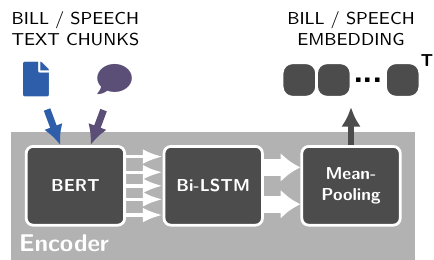}
    \caption{Overview of our Encoder: The bill/speech chunks are embedded by BERT. The Bi-LSTM computes an aggregated embedding for speeches/bills, and the mean pooling layer reduces their dimensionality.}
    \label{fig:encoder1}

\end{figure}

The aim of our Encoder is to compute textual embeddings for bills and speeches while preserving the contextual information contained in the texts and transcripts of these documents.
When developing such an encoder, we have to solve the problem that both bills and speeches have lengths exceeding the embedding capabilities of state-of-the-art language models \cite{devlin2018bert, beltagy2020longformer}. 
In our case, the average number of words for bills and speeches is $2239.43$ and $8129.23$, respectively. 
We, therefore, propose the Encoder architecture shown in \Cref{fig:encoder1} in which we split the original bills/speech documents $D$ into 512-word chunks $C_i$, i.e., $D=\{C_1, C_2, ..., C_T\}$.
Subsequently, we use BERT \cite{devlin2019} to compute embedding vectors $C_i^{bert}$ for each chunk $C_i$.
We then use a Bi-directional Long-Short-Term-Memory (Bi-LSTM) neural network \cite{hochreiter1997long} to combine the individual BERT embeddings.
The Bi-LSTM processes the BERT embeddings of a document's chunks both in a forward and a backward direction, aggregating them to two hidden states $\overrightarrow{h}_{T}$ and $\overleftarrow{h}_{T}$.
In a final step, we concatenate and mean-pool them to obtain the final document embedding $f = \left[\overrightarrow{h}_{T};\overleftarrow{h}_{T}\right]$.

Vocabulary and grammar of written and spoken language can differ considerably \citep{akinnaso1982differences, biber1991variation}.
To account for this, we train separate Encoder instances for the bill texts and speech transcripts (see \emph{Bill} and \emph{Speech Encoder} in \Cref{fig:infographic}).

\subsection{Relational Graph Convolutional Network}\label{sec:RGCN}

Our bill and speech encoders yield embeddings for all bills and speeches, respectively.
To model the \emph{relations} of legislators with these bills and speeches, we use a multi-relational heterogeneous graph $\mathcal{G} = (\mathcal{V}, \mathcal{E})$.
\begin{itemize}[label=\raisebox{.05em}{\tiny\faPlay}]
    \item $\mathcal{V} = \{S, L, B\}$ is the set of all nodes where $S$ is the set of speeches, $L$ is the set of legislators and $B$ is the set of bills.
    The bill and speech nodes are initialized with the embeddings computed by the encoders.
    Legislator nodes are initialized with a hot-one encoding of their metadata (see \Cref{sec:Data}). 
    \item $\mathcal{E}$ is the set of edges.
    All edges (${u},{v},{r}) \in \mathcal{E}$ have a source $u$, a target $v$, and a relation type $r \in \mathcal{R}$.
    The set of possible relations $\mathcal{R}= \{R_1, R_2, R_3, R_4, R_5\}$ contains:
    $R_1$ authorship of speech;
    $R_2$ citation of legislator (directed);
    $R_3$ sponsorship of bill;
    $R_4$ active cosponsorship of bill;
    $R_5$ passive cosponsorship of bill.
\end{itemize}

Based on this heterogeneous graph, we employ a two-layer RGCN \cite{schlichtkrull2018modeling}.
RGCNs are graph neural networks specifically designed to learn representations for multi-relational data.
With each layer, the RGCN iteratively updates the initial embeddings of nodes based on their neighborhood, while accounting for the type of relation with the neighbors.
This means that for each node $v \in V$, our RGCN computes its embedding $e^{(k+1)}_v$ in its convolutional layer $(k+1)$ as
\begin{equation*}
    e^{(k+1)}_{v} = \sigma\left(\sum_{r\in \mathcal{R}}\sum_{j\in \mathcal{N}^{r}_{v}} \frac{W_{r}^{(k)}e_{j}^{k}}{c_{v,r}} + W_{0}^{k}e_{v}^{k}\right),
\end{equation*}
where $\mathcal{N}^{r}_{v}$ is the set of neighbors of node $v$ connected by relation of type $r$, $\sigma$ is the activation function, $c_{v,r}$ is a normalization constant, and $W_r$ and $W_0$ denote the relation specific transformations used by the RGCN during the training.
As suggested by \citet{schlichtkrull2018modeling}, we set $c_{v,r} = |\mathcal{N}^{r}_{v}|$.
As a result, our RGCN yields holistic representations of legislators based on the speeches they give, the bills they sponsor and cosponsor, and the other legislators they cite in speeches.

\subsection{Model Training}\label{sec:model_training}

\begin{figure}
\centering
    \includegraphics[width=.5\linewidth,trim=3 0 0 0, clip]{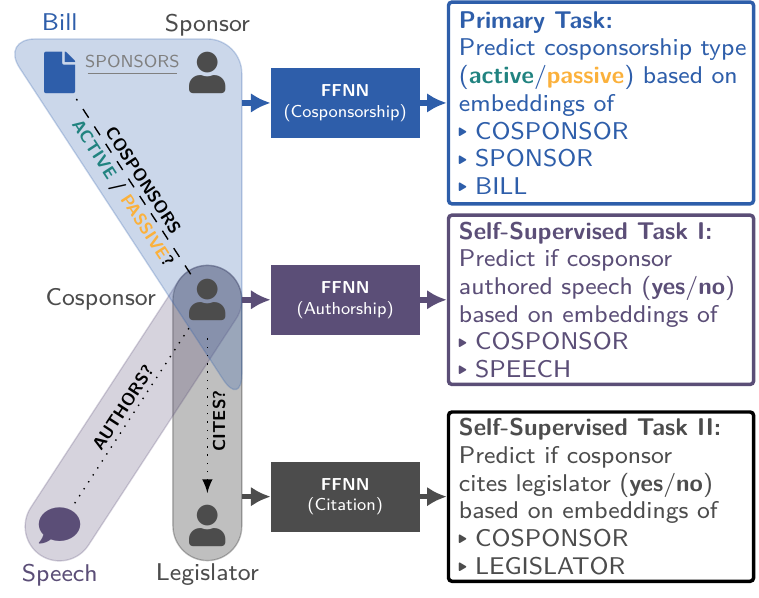}
    \caption{Overview of our classification tasks: In the primary task, we predict active and passive cosponsorship. In the self-supervised tasks, we predict if the cosponsor is the author of a speech and if the cosponsor cited another colleague in their speeches.}\label{fig:classification}
\end{figure}

We train our model by minimizing the joint loss function $\mathcal{L}_{\text{tot}}$ of three tasks:
\begin{align*}
    \mathcal{L}_{\text{tot}} = \lambda_1\mathcal{L}_{\text{cosp}} + \lambda_2\mathcal{L}_{\text{auth}} + \lambda_3\mathcal{L}_{\text{cit}},
\end{align*}
where $\lambda_1=0.8$ and $\lambda_2=\lambda_3=0.1$.
$\mathcal{L}_{\text{cosp}}$ relates to our primary task of predicting active and passive cosponsorship.
$\mathcal{L}_{\text{auth}}$ and $\mathcal{L}_{\text{cit}}$ are the losses from \emph{authorship prediction} and \emph{citation prediction}, two additional self-supervised tasks that we use to improve our model's representation of legislators. 
An overview of the three tasks, which we detail in the paragraphs below, is shown in \Cref{fig:classification}.
We provide summary statistics for training and validation data and report the results of the self-supervised tasks in \Cref{sec:training_appendix}.
We assess how the two self-supervised tasks influence our prediction performance in an ablation study (see \Cref{sec:AppendixAblation}).

\paragraph{Cosponsorship Classification}\label{sec:cosponsorship_prediction}
The primary task of our model is to predict whether a legislator's cosponsorship for a bill is \emph{active} or \emph{passive}.
Active and passive cosponsorship are mutually exclusive.
This means that a legislator $l \in L$ in the set of cosponsors $\mathcal{C}(b)$ of a bill $b \in B$, must be either an active cosponsor, $l \in \mathcal{C_{A}}(b)$, or a passive cosponsor, $l \in \mathcal{C_{P}}(b)$.
Therefore, we can formalize active/passive cosponsorship classification as computing the probability that $l$ is in the set of active cosponsors $\mathcal{C_{A}}(b)$ of bill $b$, given the bill $b$, the bill's sponsor $\mathcal{S}(b)$, and the knowledge that $l$ is a cosponsor of the bill.
\begin{align*}\label{eq:classification_task}
    p_{\mathcal{A}} = p(l \in \mathcal{C_{A}}(b) | b, \mathcal{S}(b), l \in \mathcal{C}(b))
\end{align*}
To compute $p_{\mathcal{A}}$, we concatenate the node embeddings of the legislator $l$, the bill $b$ and the bill's sponsor $\mathcal{S}(b)$.
We use concatenated embeddings as input for an FFNN with softmax which returns $p_{\mathcal{A}}$.
We use a binary cross-entropy loss to train the model for this classification task:
\begin{equation*}
    \mathcal{L}_{\text{cosp}} = - \left(y_{\mathcal{A}} \log p_{\mathcal{A}} + y_{\mathcal{P}} \log (1-p_{\mathcal{A}})\right),
\end{equation*}
where $y_{\mathcal{A}}$ and $y_{\mathcal{P}}$ are binary vectors indicating if the true cosponsorship is active or passive, respectively.

\paragraph{Authorship Prediction}
With our primary task, we aim to distinguish between active and passive cosponsorship based on the embeddings of legislators and the cosponsored bill.
To ensure that our model appropriately learns the nuances between the speeches of different legislators, we introduce our first self-supervised task, authorship prediction.
For this task, we first sample a speech $s$ every time a legislator $l$ cosponsors a bill.
To obtain an equal representation of positive and negative classes, we bias our sampling such that, with a probability of 50\%, $s$ was given by $l$.
In a binary classification task, we then use an FFNN that takes the embeddings of the cosponsor $l$ and the speech $s$ as inputs and computes the probability $p_{\text{auth}}$ that $l$ is the author of $s$. 
We evaluate the performance of our classifier using the binary cross-entropy loss $\mathcal{L}_{\text{auth}}$, where $y_{\text{auth}}$ is $1$ if legislator $l$ is the speaker of the speech $s$, is zero otherwise.
\begin{equation*}
    \mathcal{L}_{\text{auth}} = -y_{\text{auth}} \log p_{\text{auth}} - (1-y_{\text{auth}}) \log (1-p_{\text{auth}})
\end{equation*}

\paragraph{Citation Prediction}

With our second self-supervised task, we ensure that our model learns the social relationships between legislators expressed in the citations of other legislators in their speeches.
To this end, we sample a legislator $l_o$ every time a legislator $l_c$ cosponsors a bill.
We again bias our sampling such that, with a probability of 50\%, $l_c$ cites $l_o$.
We use a third FFNN, which outputs the probability $p_{cit}$ that $l_c$ cited $l_o$.
To train the model, we use again a binary cross-entropy loss $\mathcal{L}_{\text{cit}}$, where $y_{\text{cit}}$ is $1$ if $l_c$ cited $l_o$ and $0$ otherwise.
\begin{equation*}
    \mathcal{L}_{\text{cit}} = -y_{\text{cit}} \log p(y_{\text{cit}}) - (1-y_{\text{cit}}) \log (1-p(y_{\text{cit}}))
\end{equation*}

\section{Experimental Setup}\label{sec:ExperimentalSetup}

\paragraph{Data Set Splits} 
We perform a time-based splitting of our full data set for each Congress.
Specifically, we consider the first 60\% of each Congress period as training data, the subsequent 20\% as validation data, and the final 20\% as test data.
For active and passive cosponsorship classification, this yields, a total of $370,000$ training observations, and $120,000$ validation and testing samples, each.

\paragraph{Implementation Details}
We use BERT (\texttt{bert-base-uncased}) from the HugginFace library \cite{wolf2019huggingface}. 
We fine-tune our two language models (LMs) for $5$ epochs, following the indication provided by \citet{devlin2018bert}.
The dimension of the BERT embeddings is set to $768$.
We use the implementation of Bi-LSTM from PyTorch \cite{paszke2019pytorch}.
We set the hidden states dimension of the Bi-LSTM to $384$. 
Finally, the mean pooling layer at the end of the encoder outputs the initial node embeddings whose dimension is set to $128$.
To implement the RGCN, we use the DGL library \cite{wang2019dgl}.
We use $2$ layers for the RGCN as motivated by model performance (reported in \Cref{sec:training_appendix}).
The  hidden layer sizes of the two convolutional layers are $128$ and $64$, respectively.
Additionally, we use three different one-layer FFNNs with a softmax activation function for our three tasks (cosponsorship, author and citation prediction). 
These FFNNs have dimensions $192$, $128$, and $128$, respectively.
To train the model, we use AdamW \cite{loshchilov2017decoupled} as optimizer.
We tested the following learning rates for the AdamW: \{$10^{-1}$, $10^{-2}$, $10^{-3}$, $10^{-4}$\}. 
We obtain the best results with a learning rate of $10^{-4}$.
Additionally, we train our model with a batch size of $64$.
We add dropout regularization \cite{srivastava2014dropout}  and early stopping to prevent the model from over-fitting.
We stop the training after 8 epochs.

\paragraph{Baselines}
We test our model against seven baselines (\textit{B1} to \textit{B7}) which predict active and passive cosponsorship based on different representations of the bill, its sponsor, and the cosponsor.
The first two baselines differ only in the way legislators are represented.
In \emph{B1 Ideology}, legislators are represented by their ideology scores computed according to \citet{gerrish2011predicting}.
Instead, \textit{B2 Metadata} represents legislators using their metadata introduced in \Cref{sec:Data}. 
In both cases, bills are captured by their topic (e.g., healthcare) and the predictions are made using a Random-Forest-Classifier.
Analogous to \Cref{sec:model_training}, all other baselines make predictions using an FFNN.
To this end, \textit{B3 GloVe} represents each bill based on the top 200 unigrams they contain and legislators using the top 200 unigrams in their speeches using \textsc{GloVe-840B-300D} \cite[][]{Pennington2014} pre-trained word vectors.
\textit{B4 Encoder} instead obtains bill and speech representations using our Encoder introduced in \Cref{sec:Encoder}.
To obtain representations for legislators, we then average the representations or their speeches.
Baseline \textit{B5 Encoder + Metadata} uses the identical approach, but extends legislator representations using their corresponding metadata.
Our final two baseline models operate on the multi-relational heterogeneous graph introduced in \Cref{sec:RGCN}.
As these baselines do not consider textual information from our Encoder, the representations for legislators and bills are initialized randomly, and the speech nodes are excluded.
Based on this graph, \textit{B6 GCN} learns representations for legislators and bills using a Graph Convolution Network (GCN) \citep{zhang2019graph}.
Instead, \textit{B7 RGCN} uses an RGCN accounts for the multiple types of relations existing in the data.

\section{Results}

\paragraph{Model Performance} We used the model specified in \Cref{sec:Methods} and compare it to the baselines introduced in \Cref{sec:ExperimentalSetup} for our primary task of active and passive cosponsorship prediction.
Summarizing our findings, our model yields a high prediction performance with an F1-score of 0.88.
This was only possible because we incorporate contextual language and relational features of legislators and information about the bills they support to predict cosponsorship decisions. 
The results reported in \Cref{tab:model_performance} demonstrate that our model outperforms all seven baselines.
Our model has better performance than the \textit{B1 Ideology} and the \textit{B2 Metadata}, which relies on simple legislator characteristics, of $14\%$ and $15\%$ respectively. 
This means that simple characteristics of legislators cannot sufficiently explain their cosponsorship behavior.
Adding contextual information, \textit{B4 Encoder} increases the prediction performance over \textit{B1} and \textit{B2} by roughly the $10\%$.  
This points to a topical alignment between the speeches of legislators and the bills they cosponsor.
By combining the RGCN with the \textit{Encoder}, our model utilizes both language and relational information (citation, authorship and cosponsorship), resulting in an F1-score of 0.88.
To conclude, the combination of textual and relational information proves to be key for an accurate prediction of cosponsorship decisions. 

\begin{table*}[t]
\small
\begin{tabularx}{\linewidth}{cCCCCCCCC}
    \toprule
    Congress & \multicolumn{1}{c}{Ideology} & \multicolumn{1}{c}{Metadata} & \multicolumn{1}{c}{GloVe} & \multicolumn{1}{c}{Encoder} & \multicolumn{1}{c}{\thead{Encoder +\\ Metadata}} & \multicolumn{1}{c}{GCN} & \multicolumn{1}{c}{RGCN} & \thead{\textbf{Encoder +}\\ \textbf{RGCN}} \\ \midrule
    112      & 0.735\footnotesize{$\pm$0.02} & 0.739\footnotesize{$\pm$0.04} & 0.773\footnotesize{$\pm$0.05} & 0.832\footnotesize{$\pm$0.03} & 0.829\footnotesize{$\pm$0.05} & 0.749\footnotesize{$\pm$0.05} & 0.784 \footnotesize{$\pm$0.04} & \textbf{0.874}\footnotesize{$\pm$0.05} \\
    113      & 0.756\footnotesize{$\pm$0.04} & 0.732\footnotesize{$\pm$0.07} & 0.767\footnotesize{$\pm$0.03} & 0.839\footnotesize{$\pm$0.05} & 0.845\footnotesize{$\pm$0.06} & 0.755\footnotesize{$\pm$0.03} & 0.799 \footnotesize{$\pm$0.04} & \textbf{0.892}\footnotesize{$\pm$0.03} \\
    114      & 0.745\footnotesize{$\pm$0.03} & 0.741\footnotesize{$\pm$0.06} & 0.758\footnotesize{$\pm$0.04} & 0.843\footnotesize{$\pm$0.05} & 0.861\footnotesize{$\pm$0.06} & 0.763\footnotesize{$\pm$0.04} & 0.801 \footnotesize{$\pm$0.03} & \textbf{0.882}\footnotesize{$\pm$0.04} \\
    115      & 0.751\footnotesize{$\pm$0.03} & 0.726\footnotesize{$\pm$0.05} & 0.777\footnotesize{$\pm$0.04} & 0.846\footnotesize{$\pm$0.02} & 0.853\footnotesize{$\pm$0.04} & 0.792\footnotesize{$\pm$0.05} & 0.816 \footnotesize{$\pm$0.05} & \textbf{0.889}\footnotesize{$\pm$0.04} \\[1mm]
    Avg      & 0.746\footnotesize{$\pm$0.03} & 0.734\footnotesize{$\pm$0.05} & 0.768\footnotesize{$\pm$0.04} & 0.840\footnotesize{$\pm$0.03} & 0.847\footnotesize{$\pm$0.05} & 0.765\footnotesize{$\pm$0.04} & 0.800 \footnotesize{$\pm$0.05} & \textbf{0.884}\scriptsize{$\pm$0.04} \\ \bottomrule
    \end{tabularx}
\caption{F1-score ($\pm$s.d.) for our model (bold) and baselines for active and passive cosponsorship classification.}\label{tab:model_performance}
\end{table*}

\begin{figure}[t]
    \small
    \centering
    \begin{tikzpicture}
    \node at (0,0) {
        \includegraphics[width=.5\linewidth]{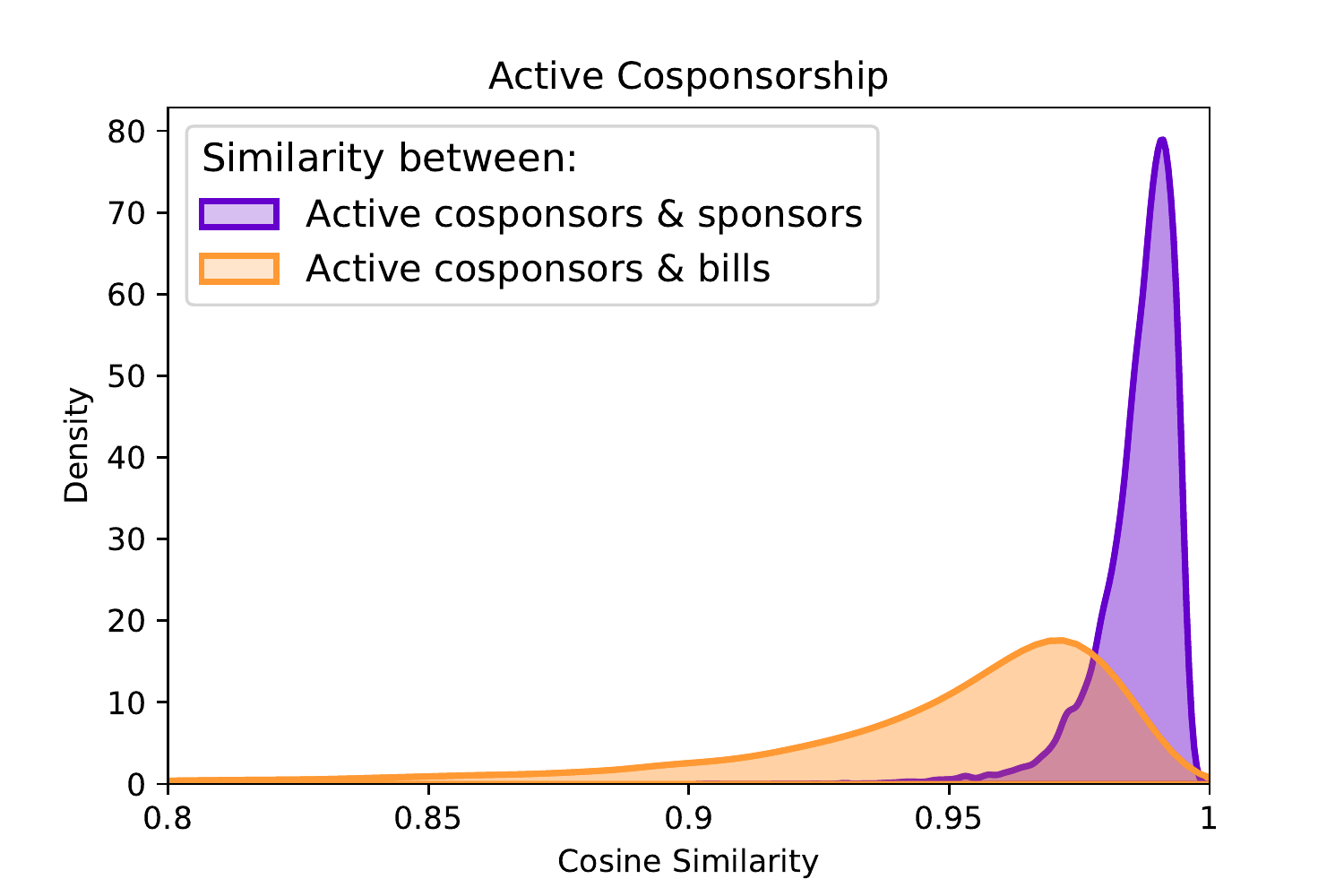}
    };
    \node at (8,0) {
        \includegraphics[width=.5\linewidth]{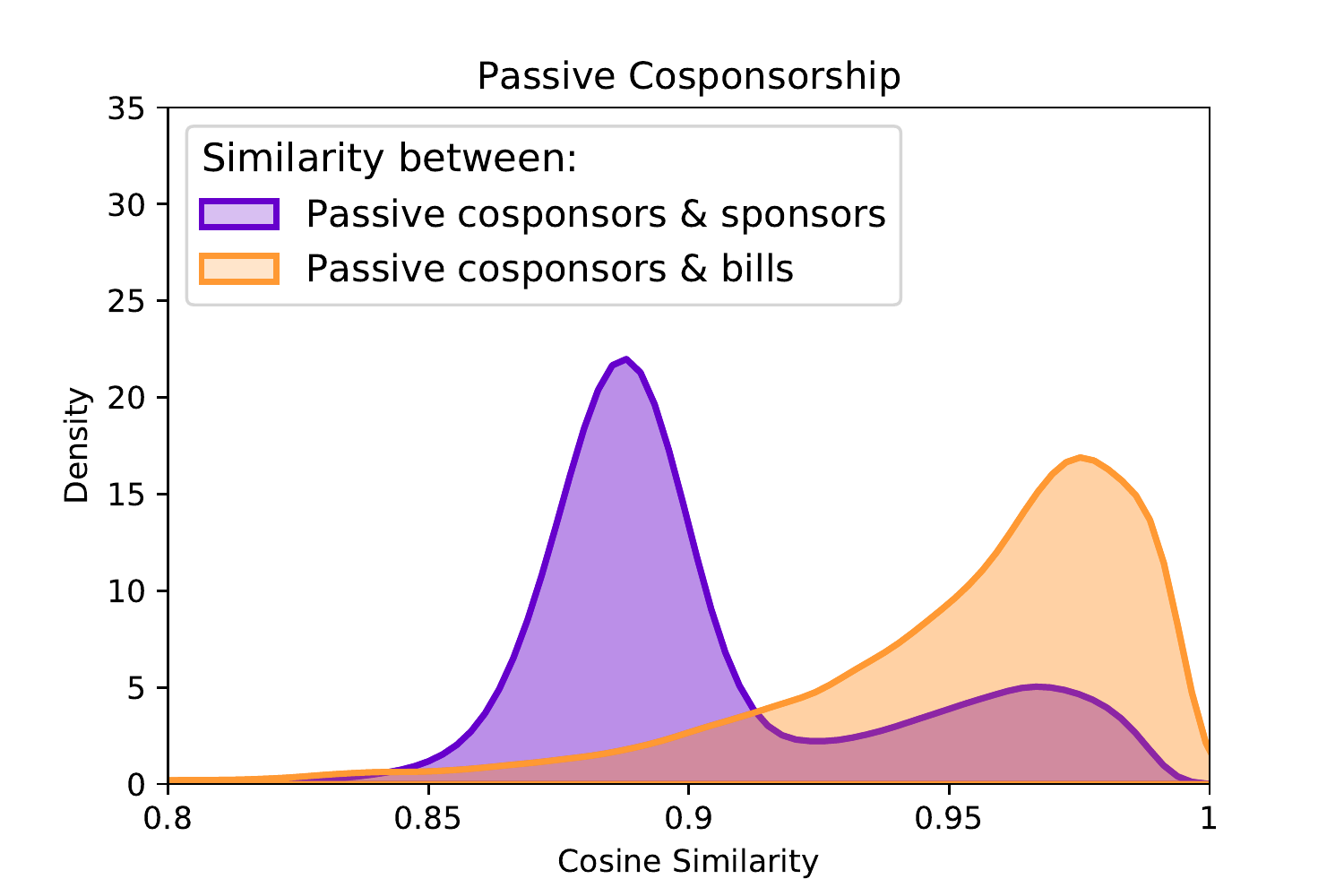}
    };
    \node at (-3.4,2.5) {\bfseries (a)};
    \node at (4.6,2.5) {\bfseries (b)};
    \end{tikzpicture}
    \caption{Density of cosine similarity between cosponsor representations and sponsor {\textcolor{densityP}{\footnotesize\faSquare}} or bill {\textcolor{densityO}{\footnotesize\faSquare}} representation. Panel (a)  shows active cosponsors. Panel (b) shows passive cosponsors.}
    \label{fig:three graphs}
\end{figure}

\paragraph{Active vs. Passive Cosponsorship}
Our model learns representations for both legislators and bills in order to predict active and passive cosponsorship.
\Cref{fig:three graphs}a illustrates that representations of \emph{active} cosponsors of a bill have a higher average cosine similarity with the representation of the \emph{sponsor} of the bill. 
This means that active cosponsorship is used to back a person, i.e., the sponsor.
Representations of \emph{passive} cosponsors, on the other hand, have a higher average cosine similarity with the representations of the \emph{bills} (see \Cref{fig:three graphs}b).
To summarize our findings, we can explain the difference between active and passive consponsorship by distinguishing between two different motivations, namely backing political colleagues or backing a bill's content. 
As such, information about active cosponsorship can provide further insights into political alliances, whereas information about passive cosponsorship can be useful in studying agenda setting and campaigning. 

\begin{figure}[t]
    \centering
    \scriptsize
    \includegraphics[trim=0 25 0 0, clip, width=.6\textwidth]{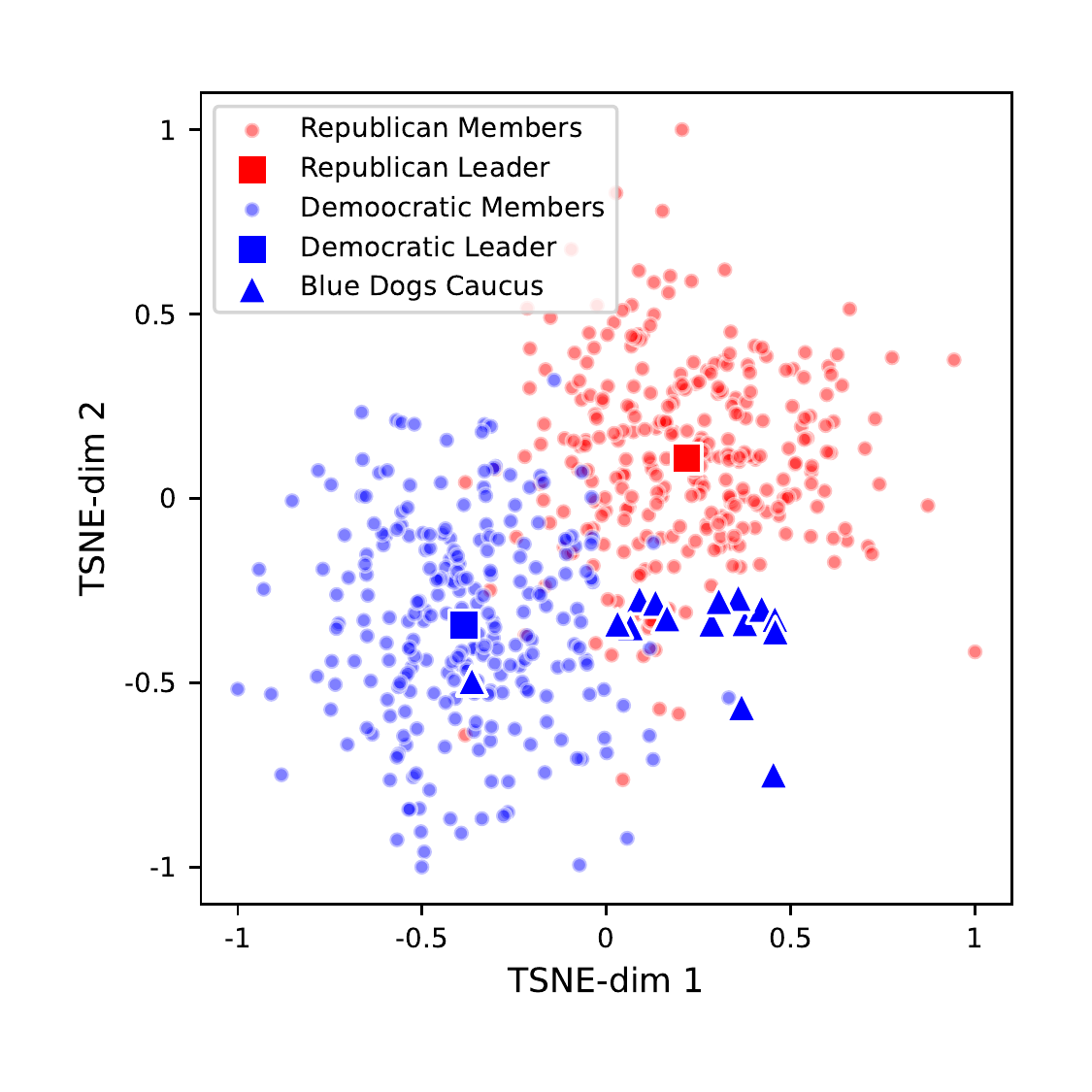} 
    \caption{2D projection of the legislator representations.
      As shown, our representation of Legislators splits them correctly along party lines ({\textcolor{partyB}{\footnotesize\faSquare}}, {\textcolor{partyR}{\footnotesize\faSquare}}).
      Party leaders are found in the center of their respective party clusters.
      We also find that members of the ``Blue Dogs Caucus'' are correctly positioned between the two parties. 
    }
    \label{fig:ideology}
\end{figure}

\paragraph{Interpretation of Legislator Representations}\label{sec:ideology}

Our legislator representations can be further used to study other legislative activities, such as voting.
We can interpret 
the representations of legislators as a proxy of their ideology
(similar to \cite{Kraft2016}).  
In \Cref{fig:ideology} we plot a two-dimensional projection (using TSNE, \cite{van2008visualizing}) of our legislator representations.
We find a clear split between Republican and Democrat legislators. 
Interestingly, Republican and Democrat party leaders are located at the center of their respective party.
Moreover, we highlight the so-called ``Blue Dog Caucus'', the group of conservative Democrats who our representations
correctly place between  Republicans and Democrats.

In a final experiment, we use an additional FFNN with representations of legislators and bills to predict the vote of a legislator on a bill (``yea'', ``nay'').
The information contained in our representations allows us to predict roll-call votes without training, matching the performance of models specifically trained for voting prediction (see details in \Cref{tab:rcvs}).
These qualitative and quantitative results show that our legislator representations are meaningful as a proxy for ideology and can be used outside cosponsorship prediction, opening doors for future tasks in the study of legislative behavior.

\section{Related Work}

The analysis of cosponsorship decisions has been widely studied by experts of political science, e.g., \cite{campbell1982cosponsoring, krehbiel1995cosponsors, mayhew2004congress}.
Research on cosponsorship often focuses on three aspects: the agenda-setting dynamics of bill introductions and cosponsorship \cite{koger2003position, kessler1996dynamics}, how cosponsorship affects bill passage \cite{wilson1997cosponsorship, browne1985multiple, woon2008bill, sciarini2021influence, Dockendorff2021cosponsorship}, and alliances between legislators \cite{fowler2006connecting, kirkland2011relational, kirkland2014measurement, lee2017time, brandenberger2018trading, brandenberger2022comparing}.

Despite political science research directly linking cosponsorship to the texts of bills and speeches in congress, cosponsorship has so far received little to no attention from the NLP community.
However, recent advances of natural language processing \cite{devlin2018bert, vaswani2017attention,zhao2019gender, russo2020control} provides tools to address numerous social science questions related to legal \cite{xu2020distinguish, valvoda2021precedent, valvoda2018using}, conflicts \cite{cui-etal-2020-edge, stoehr2021classifying} and political studies \cite{nguyen2015tea,schein2019allocative,falck2020measuring,glavavs2017unsupervised}.
Among these studies, the prediction of roll-call votes has received great attention.
For example, \citet[][]{Eidelman2018} propose a model to predict voting behavior using bill texts and sponsorship information and find that the addition of the textual information of the bill improves voting predictions drastically.
Similarly, \citet[][]{Gerrish2011} improve upon voting prediction by proposing a \textit{congress model} that proxies ideological positions of legislators by linking legislative sentiment to bill texts.
This model has been extended  to further improve predictions of roll-call votes \cite{Patil2019, Kraft2016, Karimi2019, kornilova2018party, Xiang2019, budhwar2018predicting, vafa2020text, mou2021align}.

\section{Conclusion}

In this work, we developed an Encoder+RGCN based model that learns holistic representations of legislators, accounting for the bills they sponsor and cosponsor, the speeches they give, and other legislators they cite.
This representation enabled us to predict the type of cosponsorship support legislators give to colleagues with high accuracy.
Specifically, we differentiated between \emph{active} cosponsorship, which is given before the official introduction of the bill to the Congress floor, and \emph{passive} cosponsorship, which is given afterwards.
So far, the political science literature has distinguished these forms of cosponsorship in terms of their resource-intensity \citep{fowler2006connecting} and their alliance formation dynamics \citep{brandenberger2018trading}.
However, we showed that legislators in the U.S. Congress use active and passive cosponsorship for two fundamentally different aims: active cosponsorship is used to back a colleague and passive cosponsorship serves to back a bills' agenda.

Studying the transferability of our representations to other legislative activities, we showed that the resulting legislator embeddings can be used to proxy their ideological positions.
Specifically, our representations separate legislators, matching not only their party affiliation but even their caucus membership.
Finally, in an application of zero-shot learning, we showed that our representations match task-specific state-of-the-art methods when predicting the outcomes of roll-call votes \emph{without} requiring any additional training.
Hence, our legislator representations are interpretable and generalize well to unseen tasks.

Our results have important implications for both the study of cosponsorship and future studies of U.S. legislative activities. 
For cosponsorship, when aiming to study the relations between legislators, data on \emph{active} cosponsorship should be used.
In turn, to study agenda support among legislators, the information contained in \emph{passive} cosponsorship is most meaningful.
In future research, our holistic representations of U.S. legislators allow for deeper insights into how ideology affects alliance formation, agenda setting and political influencing.

\clearpage
\appendix

\section{Data}\label{sec:AppendixData}

In this section, we decide to provide additional information about our collected data.
We provide a summary statistics of our dataset in \Cref{tab:bills}

\subsection{Cosponsoring}\label{appendix:data}
In this section, we provide additional information about all the data we used.
We collected all bills that were supported by more than $10$ cosponsors.
In particular, we collected all the bills of the following categories: (i) House/Senate Resolution, (ii) House/Senate Joint Resolution, (iii) House/Senate Concurrent Resolution.

\paragraph{Active and Passive Cosponsoring}
To show that the party affiliation does not affect significantly the distribution of active and passive labels, we provide in \Cref{appendix_fig:active_passive} an analysis of the distribution of the two labels. 
We notice that there is a higher tendency of Republicans to cosponsor both actively and passively.
\begin{figure*}[!htb]

    \minipage{0.5\linewidth}
      \includegraphics[width=\linewidth]{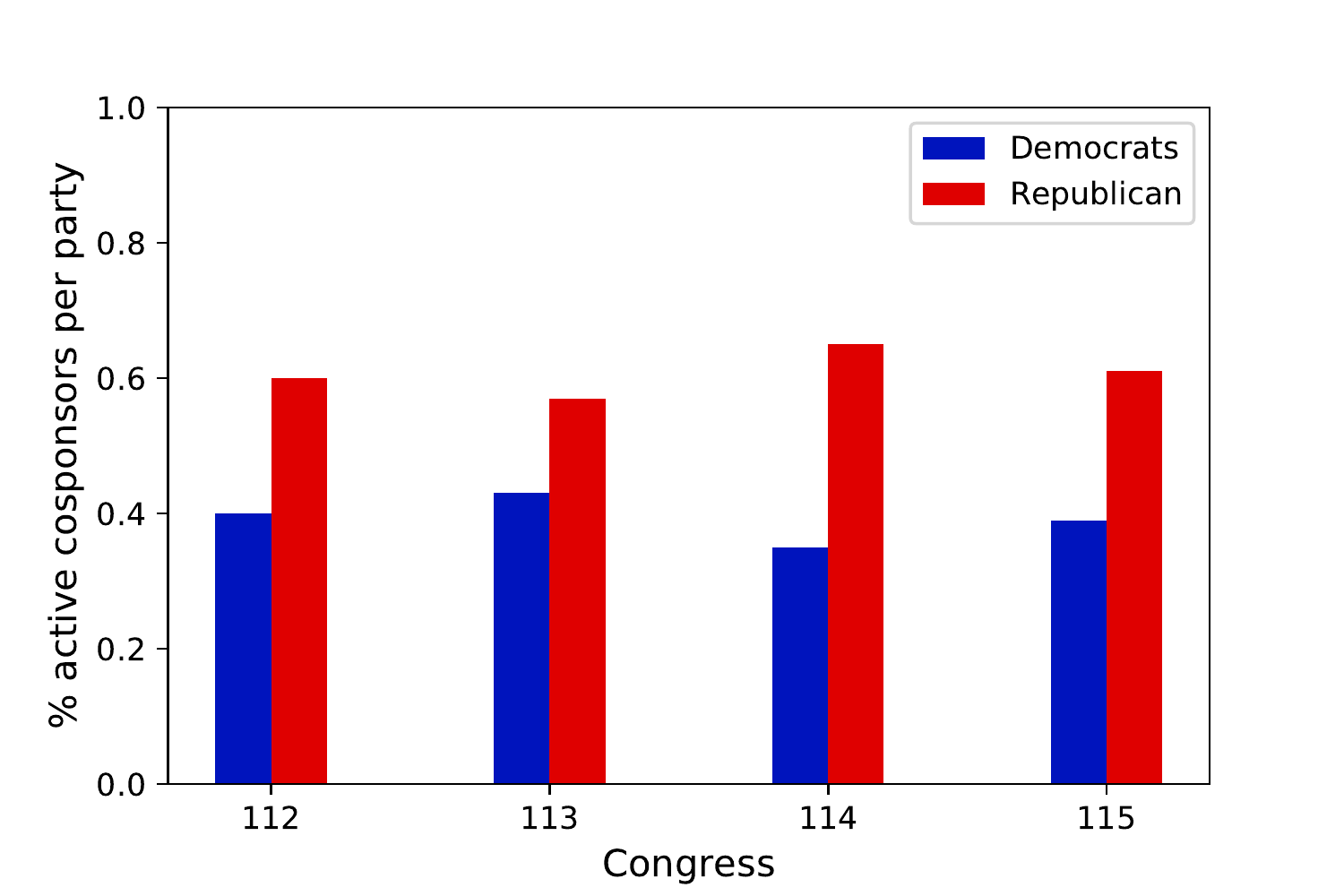}
    \endminipage\hfill
    \minipage{0.5\linewidth}
      \includegraphics[width=\linewidth]{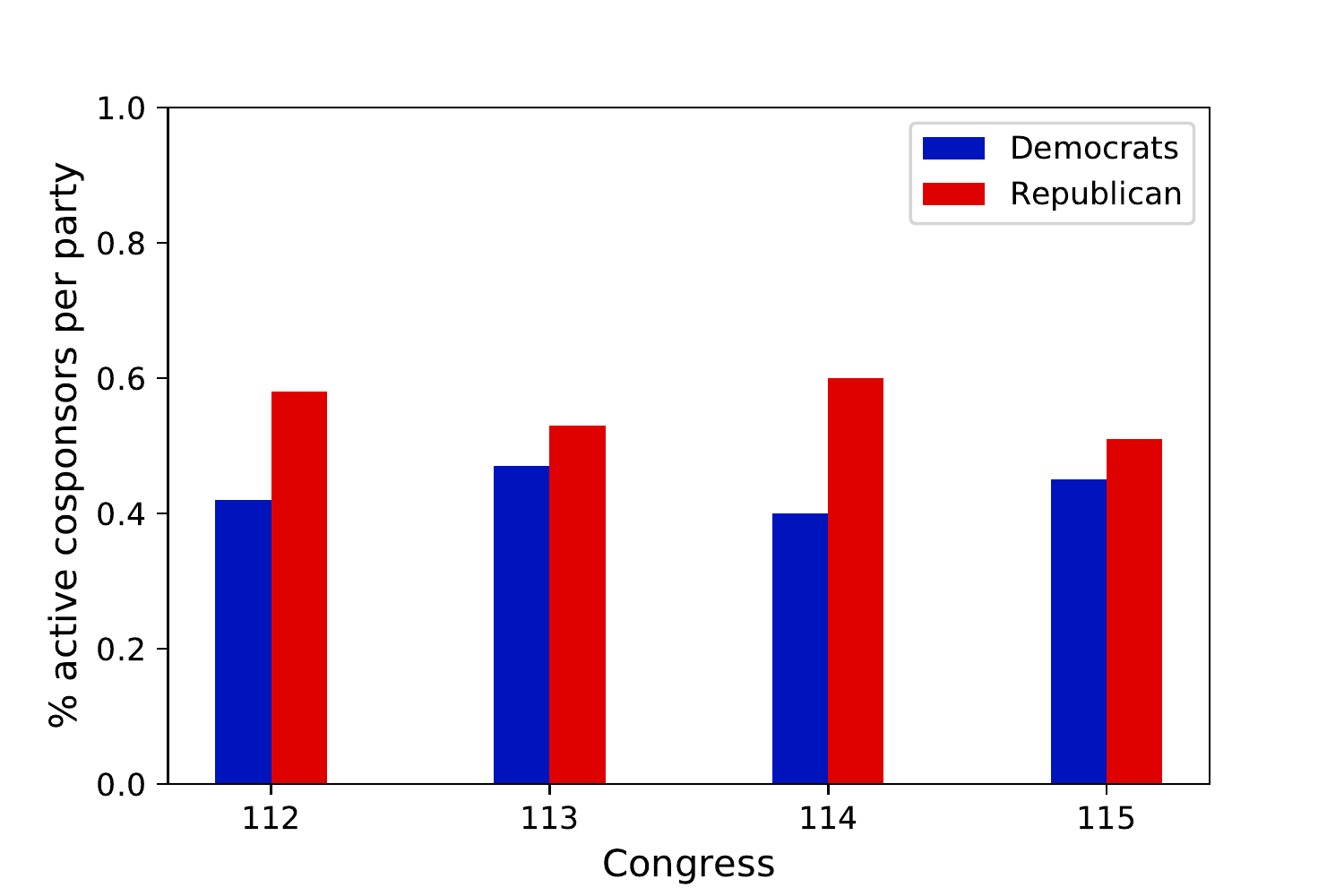}
    \endminipage
    \caption{ Distribution of active (left) and passive (right) cosponsorship across parties. }
    \label{appendix_fig:active_passive}

    \end{figure*}

Finally, in \Cref{tab:speeches}, we provide statistics about the number of speeches and how they are distributed among legislators. 
We also provide a visualization of the number of bills proposed by Republicans and Democrats during the four Congresses in \Cref{appendix_fig:bills}.

 \begin{figure}[!b]
    \centering
    \includegraphics[width=.5\linewidth]{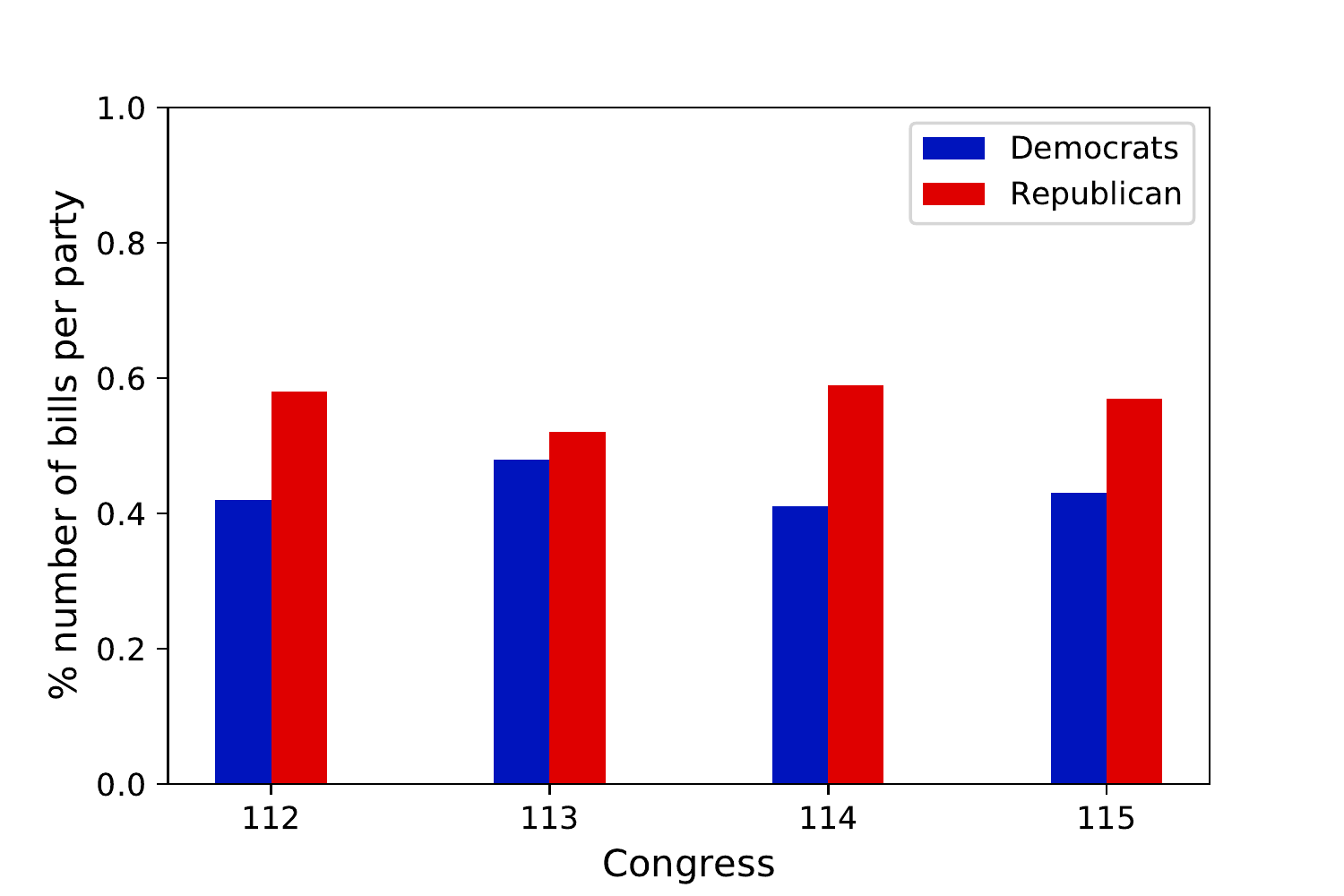}
    \caption{Distribution of the number of bills across parties for the 112th-115th congresses. }\label{appendix_fig:bills}
\end{figure}

\begin{table}[t]
    \centering
    \small
    \begin{tabularx}{.6\linewidth}{YYYY}
        \toprule
        Congress & \#Bill  & \#Active & \#Passive  \\
        \midrule
        112      & 14042                                  & 68113             & 78507    \\
        113      & 12852                                  & 63176             & 82657	\\
        114      & 14550                                   & 77746             & 82149	\\
        115      & 15754                                   & 78751             & 85308     \\
        \bottomrule	
    \end{tabularx}
        \caption{Summary statistics of bills and cosponsorship signatures.}\label{tab:bills}
   \bigskip
    \centering
    \small
    \begin{tabularx}{.6\linewidth}{CCCC}
        \toprule
        Congress & \#Speeches & \#Speeches & Speech length   \\
	 	   & (total)	  & (avg. per MP) & 	(avg. \# words)\\
        \midrule
        112      & 32189            & 60.16                     & 224.82                \\
        113      & 36623            & 68.47                       & 225.41             	\\
        114      & 30121            & 56.30                       & 218.10             	\\
        115      & 31579            & 59.02                        & 223.64                 \\
        \bottomrule	
    \end{tabularx}
    \caption{Summary statistics of congressional speeches.}\label{tab:speeches}
 \end{table}

\section{Training Results}\label{sec:training_appendix}
As discussed in \Cref{sec:model_training}, we use authorship and citation prediction as two additional self-supervised tasks to train our model. 
Here, we discuss some details about the implementation of these two tasks. 
In particular, we first discuss how the data are generated and second how the model performs on these tasks.

\begin{table}[t]
\centering
\small

\begin{tabularx}{.6\linewidth}{cCCC}
\toprule
\textbf{Model} & \textbf{Training} & \textbf{Validation} & \textbf{Test} \\ \midrule
\multicolumn{4}{l}{\textbf{Authorship Prediction}}                       \\ \midrule
Encoder        & 0.881             & 0.875               & 0.873         \\
Our model  & 0.932             & 0.921               & 0.911         \\ \midrule
\multicolumn{4}{l}{\textbf{Citation Prediction}}                                  \\ \midrule
Encoder        & 0.667             & 0.652               & 0.639         \\
Our model    & 0.699             & 0.685               & 0.665         \\ \bottomrule
\end{tabularx}
\caption{F1-scores for training, validation and testing separated for the two learning tasks.
For each task, we compare our model (Encoder + RGCN) against the encoder representations.
}
\label{tab:additional_tasks}
\end{table}

\paragraph{Authorship Prediction}
For this particular task, we first sample a speech $s$ every time a legislator $l$ cosponsor a bill. 
This speech is sampled with $30\%$ chance from the speeches that $l$ gave and with $70\%$ chance from other speeches not given by $l$.
Following this procedure, we generate our positive and negative training samples for each legislator. 
These data are split into training, validation and test sets using the same splitting scheme (60-20-20) used for the primary tasks of cosponsorship prediction (see \Cref{sec:cosponsorship_prediction}). 
We test the performance of our model on the training and validation set and compare it with the performance yield by the Encoder representations only. 
These results are shown in \Cref{tab:additional_tasks}. 

\paragraph{Citation Prediction}
Similar to the authorship prediction task, we sample a legislator $l_{o}$ every time a legislator $l_{c}$ cosponsors a bill. 
This legislator $l_{o}$ is sampled with a $50\%$ chance from the legislators that $l_{c}$ cited in their speeches.
Additionally, we substitute the name of the cited legislator $l_{o}$ with the token \texttt{<LEG>} in all the speeches of legislator $l_{c}$.
As before, we applied a 60-20-20 split to the data that we generated with this procedure. 
\Cref{tab:additional_tasks} provides the results from the performance of our model on the training and validation set and a comparison with the performance from the encoder representations only. 

\begin{figure}
    \centering
     \includegraphics[width=.6\linewidth]{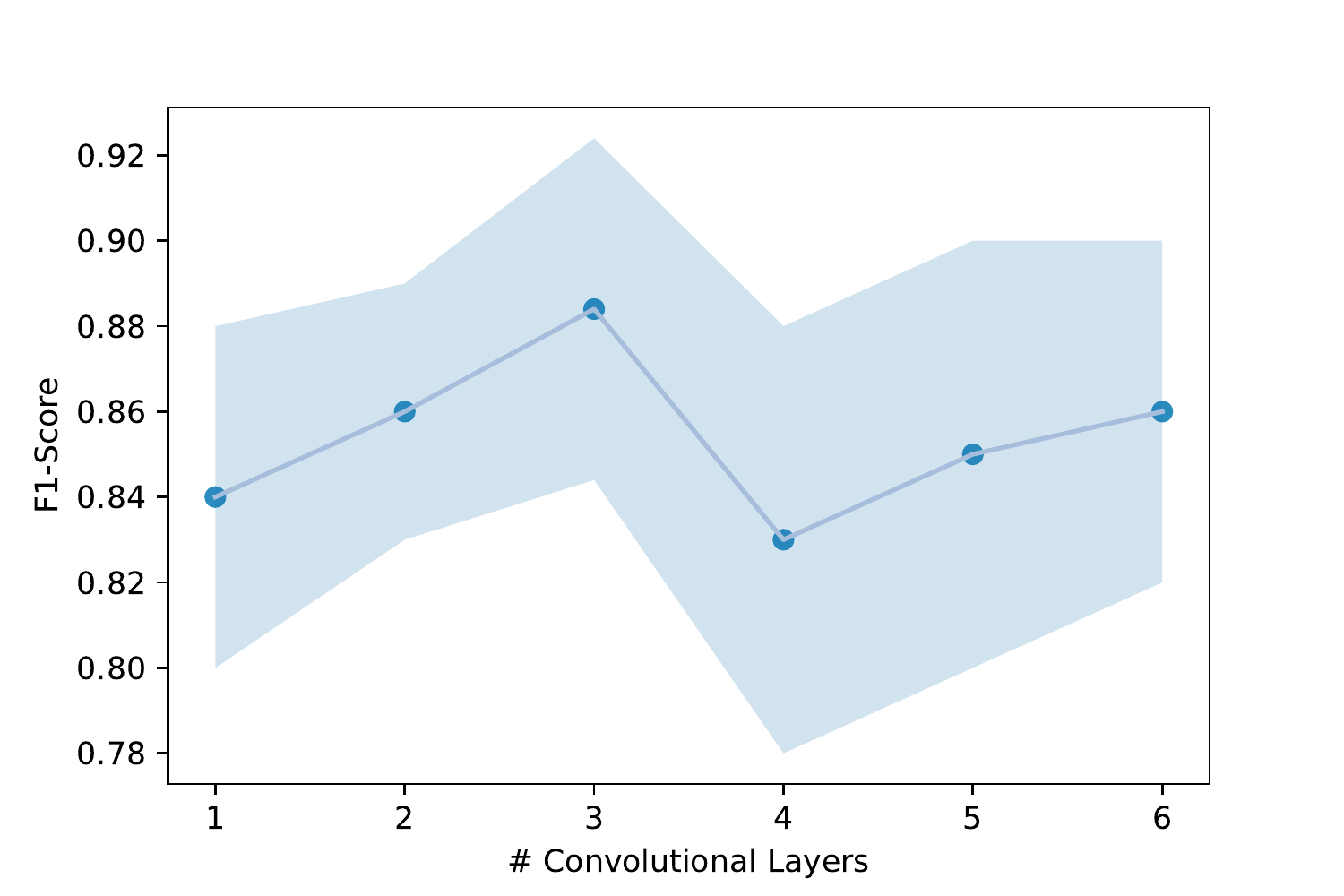}
     \caption{Model performance of our model w.r.t to the number of convolutional layers.}
     \label{fig:billprocess22}
 \end{figure}

\section{Results}
\subsection{Ablation Study}\label{sec:AppendixAblation}

We conduct an ablation study by testing how our two self-supervised tasks, authorship prediction and citation prediction, affect our overall prediction performance. 
The model trained without the two self-supervised tasks achieves a F1-score of 0.85 (see \Cref{tab:ablation_study}).
By including authorship prediction only, the  F1-score increase to 0.87.
By including citation prediction only, the same accuracy is achieved.  
Including both  tasks together, our model results in the highest F1-score of 0.88.

\begin{table}[t]
\small
\centering
\begin{tabularx}{.6\linewidth}{cCCCC}
\toprule
Congress & $\mathcal{L}_\text{cosp}$ & $\mathcal{L}_{\text{tot}}$-$\mathcal{L}_\text{auth}$ & $\mathcal{L}_{\text{tot}}$-$\mathcal{L}_\text{cit}$ & $\mathcal{L}_{\text{tot}}$ \\ \midrule
112               & 0.841       & 0.855       & 0.858       & 0.874       \\
113               & 0.847       & 0.875       & 0.871       & 0.892       \\
114               & 0.864       & 0.878       & 0.869       & 0.882       \\
115               & 0.861       & 0.871       & 0.871       & 0.889       \\ \midrule
Avg               & 0.853       & 0.870       & 0.867       & 0.884       \\ \bottomrule
\end{tabularx}
\caption{Ablation Study of the loss functions $\mathcal{L}_{\text{cosp}}$ (cosponsorship), $\mathcal{L}_{\text{auth}}$ (authorship) and $\mathcal{L}_{\text{cit}}$ (citations) for the 112th-115th congresses.}\label{tab:ablation_study} 
\end{table}

\subsection{Predicting Roll-Call Votes}

As discusssed in \Cref{sec:ideology}, we use the representations learnt by our model to predict other legislative decisions. 
In particular, we focused on the prediction of roll-call votes, which are votes expressed by a legislator on a bill (``yea'', ``nay'').
To perform this task, we train a three layer FFNN with ReLu as activation function and dropout regularization set to 0.2.
The FFNN takes as input the embeddings of the bill and of the legislator voting on that specific bill. 
To avoid leakage of information, we predict the voting decisions on bills that were not cosponsored by the legislator voting.
We compare the results of this model with four models directly trained for the task of voting predictions:
(i) \textit{Majority (Maj)} is a baseline which assumes all legislators vote yea. 
(ii) \textit{Ideal-Vectors (IV)} are multidimensional ideal vectors for legislators based on bill texts obtained following the method of  \citet{Kraft2016}. 
(iii) \textit{CNN+meta} is based on CNN and adds the percentage of sponsors of different parties as bill’s authorship information \cite{kornilova2018party}. (iv) \textit{LSTM+GCN} uses LSTM to encode legislation and applies a GCN to update representations of legislators \cite{yang2020joint}.
\Cref{tab:rcvs} shows that our model achieves an F1-score of 0.89.

\begin{table}[t]
    \small
    \centering
    \begin{tabularx}{.7\linewidth}{cCCCCC}
    \toprule
    Congress & Maj & IV & \thead{CNN+\\Meta}     & \thead{LSTM+\\GCN}              & \textbf{Ours}                  \\ \midrule
    112      & 0.787    & 0.869   & 0.885  & 0.895           & \textbf{0.907 } \\
    113      & 0.765    & 0.878   & 0.879  & 0.884          & \textbf{0.890} \\
    114      & 0.774    & 0.872   & 0.878  & \textbf{0.892} & 0.889         \\
    115      & 0.772    & 0.875   & 0.880  & 0.883           & \textbf{0.887 } \\ \midrule
    Avg      & 0.774    & 0.873   & 0.879  & 0.886          & \textbf{0.893 } \\ \bottomrule
    \end{tabularx}

        \caption{F1-scores for roll-call vote predictions. We compare our results (our representations + FFNN) to four baselines: (i) \textit{Majority (Maj)}, (ii) \textit{Ideal-Vectors (IV)} \citep{Kraft2016},  
          (iii) \textit{CNN+Meta} \cite{kornilova2018party}, and (iv) \textit{LSTM+GCN}  \cite{yang2020joint}.
        }
        \label{tab:rcvs}
    \end{table}
\end{document}